\documentclass[10pt,twocolumn,letterpaper, dvipsnames, table, xcdraw]{article}

\usepackage[dvipdfmx]{graphicx}
\usepackage{listings}
\usepackage{dirtree}

\usepackage[pagenumbers]{cvpr} 

\definecolor{cvprblue}{rgb}{0.21,0.49,0.74}
\usepackage[pagebackref,breaklinks,colorlinks,allcolors=cvprblue]{hyperref}
\usepackage[hyphens]{url}
\usepackage{hyperref}
\usepackage{breakurl}
\usepackage{siunitx}

\def\paperID{*****} 
\def\confName{CVPR}
\def\confYear{2025}

\title{AWML: An Open-Source ML-based Robotics Perception Framework to Deploy for ROS-based Autonomous Driving Software}

\author{
Satoshi Tanaka, \quad Samrat Thapa, \quad Kok Seang Tan, \quad Amadeusz Szymko, \\
Lobos Kenzo, \quad Koji Minoda, \quad Shintaro Tomie, \quad Kotaro Uetake, \\
Guolong Zhang, \quad Isamu Yamashita, \quad Takamasa Horibe, \\
TIER IV, Inc \\
}

\begin{document}
\maketitle
\begin{abstract}

In recent years, machine learning technologies have played an important role in robotics, particularly in the development of autonomous robots and self-driving vehicles.
As the industry matures, robotics frameworks like ROS 2 have been developed and provides a broad range of applications from research to production.
In this work, we introduce AWML, a framework designed to support MLOps for robotics.
AWML provides a machine learning infrastructure for autonomous driving, supporting not only the deployment of trained models to robotic systems, but also an active learning pipeline that incorporates auto-labeling, semi-auto-labeling, and data mining techniques.

\end{abstract}

\section{Introduction}
\label{sec:intro}

In recent years, machine learning technologies in computer vision have played a central role in robotics, especially in the development of autonomous robots and self-driving vehicles.
As the field has evolved, a variety of frameworks have emerged to support development efforts, with Robot Operating System (ROS) \cite{Quigley2009ROSAO} becoming a widely adopted standard for robot software, enabling applications ranging from academic research to commercial deployment.
One such framework is Autoware \cite{Autoware}, an open-source software stack built on ROS, designed to accelerate the development of autonomous driving systems.
Autoware integrates essential modules for perception, planning, and control—including sensor data processing, environment understanding, route planning, and actuation, making it easier for researchers and developers to build, test, and iterate on autonomous vehicle software.

\begin{figure}[t]
    \begin{center}
        \includegraphics[width=1.0\linewidth]{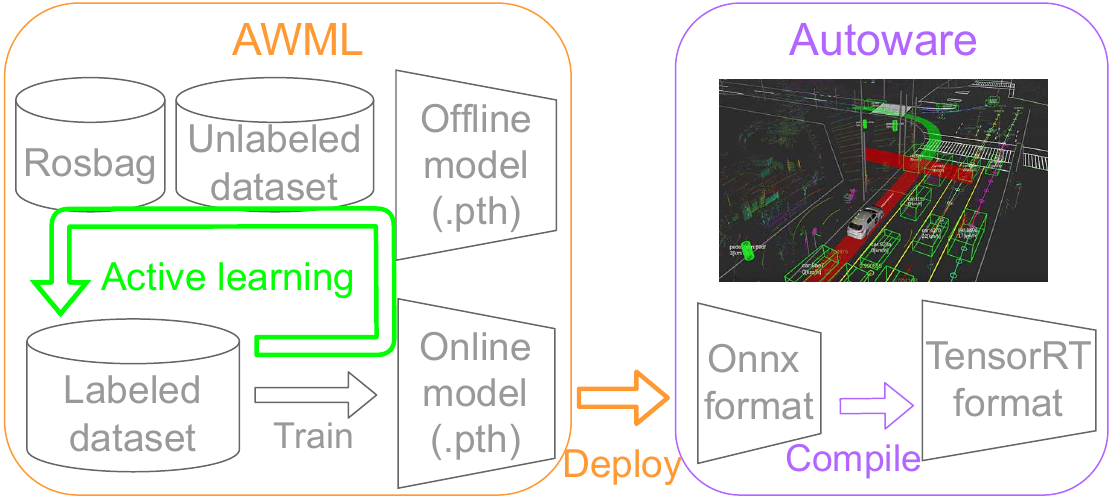}
        \caption{The framework of AWML }
        \label{AWML}
    \end{center}
\end{figure}

While academic research typically focuses on model development and evaluation in controlled environments, deploying machine learning algorithms in real-world robotics systems requires managing the entire end-to-end pipeline including data collection, training, deployment, operation, and ongoing optimization.
To address this complexity, the concept of Machine Learning Operations (MLOps) has gained traction.
MLOps practices aim to ensure that machine learning models are reproducible, stable, and maintainable throughout their lifecycle—from initial development to deployment in production systems.
However, integrating MLOps across both computer vision and robotics domains remains a significant challenge.
Many research prototypes fail to make the leap to production due to differences in tooling, programming languages, and system requirements.
For example, the robotics community often relies on real-time frameworks like ROS, written in C++ and optimized for physical deployment using accelerators such as TensorRT.
In contrast, the computer vision community largely builds on Python-based ecosystems, which provide excellent libraries and rapid prototyping capabilities, but are typically geared toward offline evaluation and lack support for real-time constraints.
Although some progress has been made at the algorithm level to close this gap, there is still a lack of unified frameworks that allow computer vision models to be developed and deployed seamlessly within robotics systems.

Active learning is also an essential component of modern MLOps workflows, especially in robotics, where data collection and annotation can be expensive and time-consuming.
To reduce labeling costs and improve model performance with limited supervision, techniques such as data mining and semi-supervised learning with pseudo-labeling are often employed.
These methods not only enhance accuracy but also enable automatic generation of new training data.
Such capabilities are particularly valuable for maintaining robust performance in complex, changing environments.
However, in practice, these techniques are still underutilized in industry, largely due to the absence of standardized interfaces and reusable frameworks that integrate them into real-world development pipelines.

In this work, we present AWML, an open-source machine learning framework tailored for ROS-based autonomous driving systems.
AWML is designed to support both the deployment of trained models into Autoware environments and the integration of active learning workflows, including auto-labeling, semi-automatic labeling, and data mining techniques, as illustrated in \cref{AWML}.
Beyond deployment, AWML provides a robotics-focused MLOps framework that helps manage the entire machine learning lifecycle, from training and version control to deployment, monitoring, and retraining.
By incorporating model versioning and fine-tuning capabilities, AWML allows developers to continuously improve models and maintain performance in real-world settings.
The key contributions of AWML to the Robotics MLOps ecosystem are as follows:

\begin{enumerate}
\item \textbf{Bridging computer vision and robotics through open-source integration}

AWML integrates with Autoware to simplify the deployment of machine learning models in autonomous driving systems.
Computer vision engineers can focus on model development using standard datasets and tools without needing to understand ROS or robotics-specific infrastructure.
At the same time, robotics engineers can handle real-machine integration using rosbag data and physical devices, without needing to manage the machine learning environment.

\item \textbf{Integrating active learning into deployment workflows}

AWML combines real-world deployment capabilities with an active learning framework.
It supports semi-supervised learning using pseudo-labeling, along with iterative data refinement through data mining techniques.
This makes it easier to efficiently generate high-quality training data from live operations.
By embedding these workflows into an MLOps context, AWML enables the continuous development and deployment of high-performance models.

\item \textbf{Connecting research tools to real-world production}

AWML uses interfaces compatible with the widely adopted MMLab ecosystem \cite{mmdetection, mmdet3d2020}, making it easier to reuse models developed in research environments.
Its loosely coupled architecture simplifies model updates and maintenance, while versioning and fine-tuning capabilities support long-term reproducibility and stability.
This allows robots to adapt more effectively to changing environments and supports safer, more reliable autonomous system deployments.
\end{enumerate}

\section{Background}
\label{sec:background}

\subsection{2D/3D Object Detection for Robotics}

2D object detection has been extensively studied, with the YOLO series becoming widely adopted in industry due to their strong real-time performance \cite{Redmon2015YouOL, Redmon2018YOLOv3AI, Cai2020YOLObileRO, bochkovskiy2020yolov4, Chen2021YouOL, Benjumea2021YOLOZIS, Wang2021YouOL, Ge2021YOLOXEY, Wang2024YOLOv9LW}.
For lightweight applications, architectures such as the MobileNet series \cite{Howard2017MobileNetsEC, Sandler2018MobileNetV2IR, Howard2019SearchingFM} and the EfficientNet series \cite{Tan2019EfficientNetRM, Tan2021EfficientNetV2SM} have also been actively explored.

3D object detection has been extensively developed based on LiDAR point cloud data, with many influential models proposed in recent years \cite{Qi2016PointNetDL, Qi2017, Zhou2017VoxelNetEL, Yan2018SECONDSE, Yang2018PIXORR3, Shi2018PointRCNN3O, Lang2018PointPillarsFE, Shi2019PVRCNNPF, Yang20203DSSDP3, Chen20203DPC, mao20233dobjectdetectionautonomous}.
Among them, CenterPoint \cite{yin2021center} is widely adopted in industry for its strong balance between detection accuracy and inference speed.
In parallel, camera-LiDAR fusion models for 3D object detection have also been actively explored \cite{Chen2016Multiview3O, Qi2018, Vora2019PointPaintingSF, wang2021pointaugmenting, Drews2022DeepFusionAR, Yingwei2022, Huang2022MultimodalSF, Li2022DelvingIT, Liang2022BEVFusionAS, Huang2024DL, li2024fully}.
TransFusion \cite{Bai2022TransFusionRL} is an attention-based architecture that enhances detection performance by fusing multi-modal features, though LiDAR-only models still often outperform it in terms of accuracy and robustness.
BEVFusion \cite{liu2022bevfusion} proposes a unified bird's-eye view (BEV) representation from multiple sensor modalities and is commonly used as a baseline for camera-LiDAR fusion studies.
In addition, camera-only 3D object detection has also been actively pursued in the computer vision community, with models such as YOLO3D \cite{Ali2018YOLO3DER} and FCOS3D \cite{Tian2019FCOSFC} demonstrating promising results despite inherent depth estimation challenges.
Lift-Splat-Shoot (LSS) \cite{Philion2020} is a widely used method in camera-based 3D object detection and camera-LiDAR fusion. It lifts 2D image features into a 3D BEV grid, serving as a key component for spatial reasoning in monocular and multi-camera systems.
In recent years, multi-camera 3D object detection has received increasing attention, with a number of models proposed to improve BEV-based detection from multiple views \cite{Huang2021BEVDetHM, li2022bevformer, Yang2022BEVFormerVA, Li2022BEVDepthAO, Zhang2022BEVerseUP, JiangLLWJWHZ24}.
StreamPETR \cite{Wang2023ExploringOT} explores object-centric temporal modeling to enhance the efficiency and temporal consistency of multi-view 3D detection systems.

Since annotation for 3D object detection is costly and time-consuming, active learning has emerged as a promising technique to reduce labeling effort while maintaining high model performance.
By selecting the most uncertain or diverse samples from a large pool of unlabeled data, active learning enables models to learn effectively with fewer labeled examples.
Several approaches have been proposed to apply active learning to 3D object detection.
For example, \cite{Jiang2022ImprovingTI} introduces a method that addresses class imbalance by mining rare examples, thereby improving detection performance on underrepresented categories.
CRB \cite{Luo2023ExploringA3} explores active learning strategies aimed at enhancing model generalization by prioritizing the labeling of key informative samples.

\subsection{Software Framework for Robotics Perception}

\begin{table*}[t]
  \centering
  \caption{
    Comparison of different 3D object detection frameworks and deployment setups.
    ``Robo. framework'' represents the framework of robotics.
    ``AL'' represents the framework of active learning.
    ``Deploy.''represents the framework of deployment for hardware optimization (like usage of TensorRT).
  }
  \begin{tabular}{l|c|c|c|l|l|l|}
  \textbf{Name}  & \textbf{Input} & \textbf{ML framework} & \textbf{Robo. framework} & \textbf{AL} & \textbf{Deploy.} \\ \hline
  YOLOv4 \cite{bochkovskiy2020yolov4, YOLOv4-Darknet-TensorRT} & 2D    &  &  \cellcolor{blue!20}ROS & & \cellcolor{blue!20}TensorRT \\
  YOLOX \cite{Ge2021YOLOXEY, Autoware-Universe} & 2D    &  &  \cellcolor{blue!20}ROS2 & & \cellcolor{blue!20}TensorRT \\
  mmdetection\_ros \cite{mmdetection-ros} & 2D & \cellcolor{blue!20}mmdetection    & \cellcolor{blue!20}ROS   &  &     \\
  ros2\_openvino\_toolkit \cite{ros2-openvino-toolkit}  & 2D    & \cellcolor{blue!20}OpenVino         & \cellcolor{blue!20}ROS2 &     &     \\
  CRB \cite{Luo2023ExploringA3} & 3D    & \cellcolor{blue!20}OpenPCDet &  & \cellcolor{blue!20}Yes &  \\
  BEVFormer \cite{li2022bevformer, BEVFormerTensorrt} & 3D    & \cellcolor{blue!20}mmdetection3d &  & & \cellcolor{blue!20}TensorRT  \\
  NVIDIA-AI-IOT \cite{NVIDIAAIIOT} & 3D    & \cellcolor{blue!20}mmdetection3d &  &  & \cellcolor{blue!20}TensorRT \\
  ROS-based-3D-detection-Tracking \cite{ROS-based-3D-detection-Tracking} & 3D & \cellcolor{blue!20}mmdetection3d & \cellcolor{blue!20}ROS2  &     &     \\
  apollo + Paddle \cite{Apollo}   & 2D/3D & \cellcolor{blue!20}Paddle libraries & \cellcolor{blue!20}Apollo Cyber RT  &       & \cellcolor{blue!20}TensorRT \\ \hline
  Ours (Autoware + AWML)  & 2D/3D & \cellcolor{blue!20}mmdetection 2d/3d  & \cellcolor{blue!20}ROS2    & \cellcolor{blue!20}Yes & \cellcolor{blue!20}TensorRT \\ \hline
  \end{tabular}
  \label{framework}
\end{table*}

Several open-source machine learning frameworks have been developed to support 2D and 3D object detection across a variety of hardware and application domains.
Paddle \cite{paddle}, developed by Baidu, is a deep learning platform designed for efficient model training, deployment, and scaling.
OpenVINO \cite{openvino}, a toolkit developed by Intel, focuses on accelerating deep learning inference on Intel hardware, supporting deployment on CPUs, GPUs, and specialized devices such as VPUs.
For 2D object detection, MMDetection \cite{mmdetection} is a widely adopted PyTorch-based framework offering a broad range of pre-trained models and training tools.
For 3D tasks, OpenPCDet \cite{openpcdet2020} is a specialized framework optimized for LiDAR-based 3D object detection.
MMDetection3D \cite{mmdet3d2020} extends MMDetection to support 3D object detection using both LiDAR and multi-modal sensor fusion.
It provides implementations of popular 3D models and enables flexible integration of data from LiDAR, cameras, and depth sensors to improve 3D perception performance.
These frameworks serve as foundational building blocks for developing and deploying perception systems in autonomous driving, robotics, and other real-world applications.

Several open-source frameworks have been developed to support robotics software development by providing modular architectures, communication tools, and control interfaces.
The Robot Operating System (ROS) \cite{Quigley2009ROSAO} is one of the most widely adopted frameworks in both academia and industry.
It simplifies the integration of modular components and offers essential tools for robot control, sensor integration, and inter-process communication.
To meet the demands of more scalable and real-time applications, its next-generation version, ROS2 \cite{ROS2}, was introduced.
ROS2 improves on ROS by offering enhanced performance, support for real-time systems, and better middleware abstraction for distributed robotics applications.
In the context of autonomous driving, Autoware \cite{Autoware} is a full-stack open-source platform built on ROS and ROS2.
It integrates modules for perception, planning, control, and simulation, and is widely used for research and development in autonomous vehicles.
Similarly, Apollo \cite{Apollo}, developed by Baidu, is an open-source autonomous driving platform based on Apollo Cyber RT, a high-performance, real-time framework for handling data pipelines and system modularity.
Apollo provides a comprehensive set of software modules covering perception, localization, planning, and control, enabling scalable and safe deployment of autonomous driving systems.

As an open-source machine learning framework for robotics perception, we summarize relevant existing toolchains and deployment setups in \cref{framework}.
For 2D object detection, YOLOv4 \cite{bochkovskiy2020yolov4} includes a TensorRT implementation, and a ROS integration is provided in \cite{YOLOv4-Darknet-TensorRT}.
Similarly, YOLOX \cite{Ge2021YOLOXEY} offers TensorRT support, with a ROS2-compatible wrapper available in Autoware \cite{Autoware-Universe}.
The mmdetection-ros package \cite{mmdetection-ros} integrates MMDetection-based 2D models into ROS environments.
In addition, the ros2\_openvino\_toolkit \cite{ros2-openvino-toolkit} supports the deployment of OpenVINO-optimized models in ROS2-based systems.
For 3D object detection, BEVFormer \cite{li2022bevformer} has a TensorRT deployment example available through \cite{BEVFormerTensorrt}.
NVIDIA-AI-IOT \cite{NVIDIAAIIOT} provides a real-time 3D perception framework with TensorRT acceleration, supporting models such as PointPillars \cite{Lang2018PointPillarsFE}, CenterPoint \cite{yin2021center}, TransFusion \cite{Bai2022TransFusionRL}, and BEVFusion \cite{liu2022bevfusion}.
Similarly, DL4AGX \cite{DL4AGX} offers a high-performance deployment stack for 3D object detection with support for BEVFormer, BEVDepth \cite{Wang2023ExploringOT}, and PillarNet \cite{JiangLLWJWHZ24}.
The ROS-based-3D-detection-Tracking project \cite{ROS-based-3D-detection-Tracking} provides a ROS2-compatible implementation that combines 3D detection and tracking, enabling direct integration into robotic systems.
For full-stack solutions, Apollo \cite{Apollo} includes both 2D and 3D object detection modules built with Paddle libraries and optimized using TensorRT, supporting comprehensive autonomous driving workflows.

\section{Method}

In this paper, we present the design of AWML and outline our approach to robotics-oriented MLOps.
Our goal is to advance MLOps practices in both machine learning and robotics, enabling the development of adaptive, scalable robotic systems that can respond quickly to new environments while maintaining high performance.

\subsection{Pipeline of AWML}
\label{method-pipeline}

\begin{figure}[t]
    \begin{center}
        \includegraphics[width=0.99\linewidth]{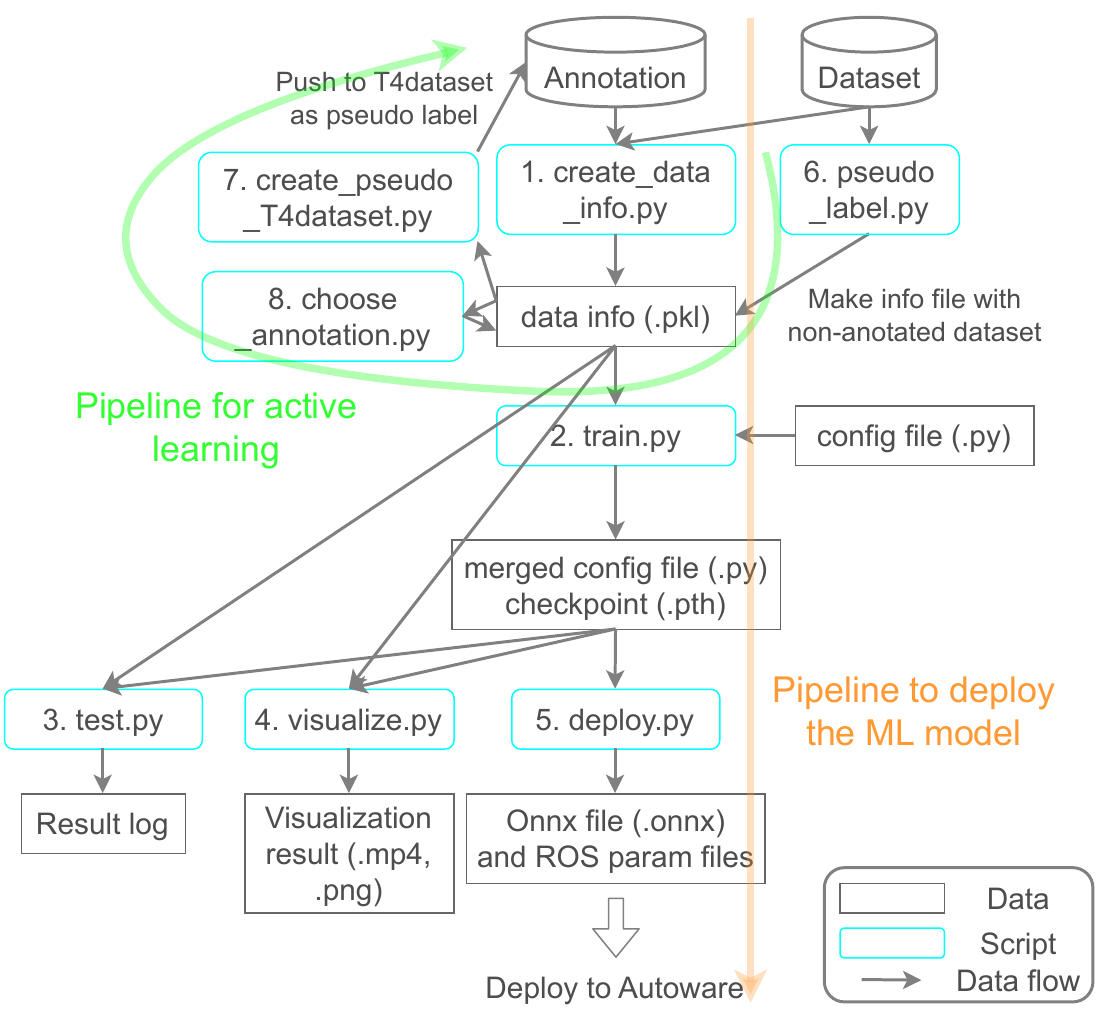}
        \caption{The pipeline of AWML.}
        \label{pipeline}
    \end{center}
\end{figure}

We build AWML on top of the MMDetection \cite{mmdetection} framework for 2D object detection and MMDetection3D \cite{mmdet3d2020} for 3D object detection.
An overview of the system architecture is shown in \cref{pipeline}.
For deployment into Autoware, the workflow consists of several key stages: training, evaluation, and deployment.
The process begins with the generation of an info file using annotated data via the create\_data\_info.py script (1).
Then, the machine learning (ML) model is trained using train.py (2), evaluated with test.py (3), and visualized using visualize.py (4).
Once the model is ready for deployment, it is exported to ONNX format using the deploy.py script (5), allowing integration into the Autoware runtime environment.

As part of AWML's active learning framework, the following components—while not directly used in Autoware deployment—play a key role in enhancing the performance of online machine learning models by leveraging offline training.
The process begins by generating an info file from an unlabeled T4dataset for both 2D and 3D data using the pseudo\_label.py script (6).
This is followed by the creation of a pseudo-labeled T4dataset via create\_pseudo\_t4dataset.py (7), and then refinement of annotations through choose\_annotation.py (8), which selects confident predictions from the raw pseudo labels.
The resulting .pkl info file is used in AWML's auto-labeling pipeline—particularly in tools such as scene\_selector and pseudo\_label, where confidence thresholds can be adjusted based on offline model predictions.
This pipeline enables the application of active learning through a 3D detection model using point cloud data, supporting more efficient and adaptive model training under limited annotation conditions.

\subsection{Type of ML Model}
\label{method-model-type}

\begin{figure*}[t]
    \begin{center}
        \includegraphics[width=0.90\linewidth]{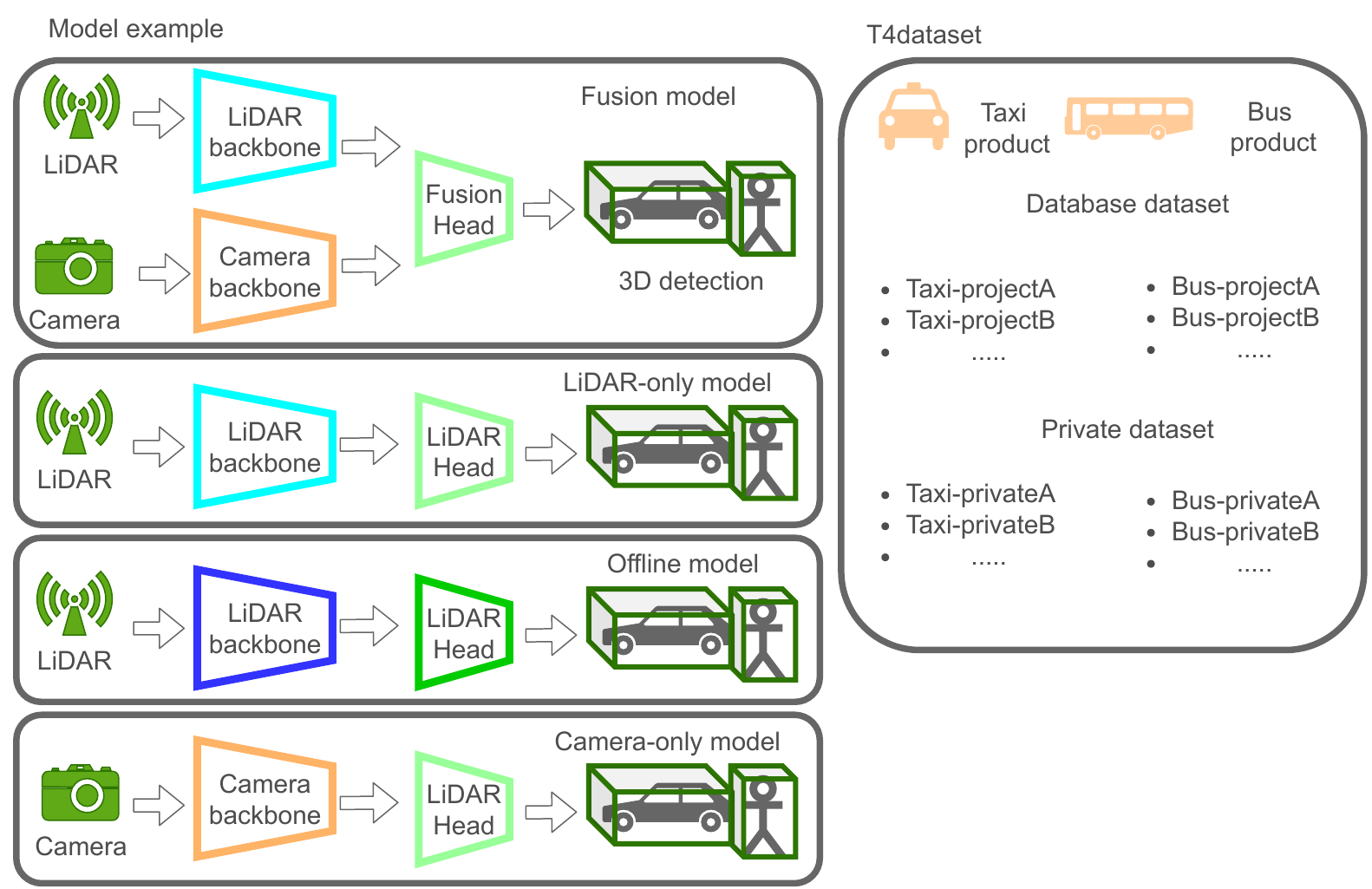}
        \caption{Model examples and T4dataset.}
        \label{model-examples}
    \end{center}
\end{figure*}

\begin{figure*}[t]
    \begin{center}
        \includegraphics[width=0.95\linewidth]{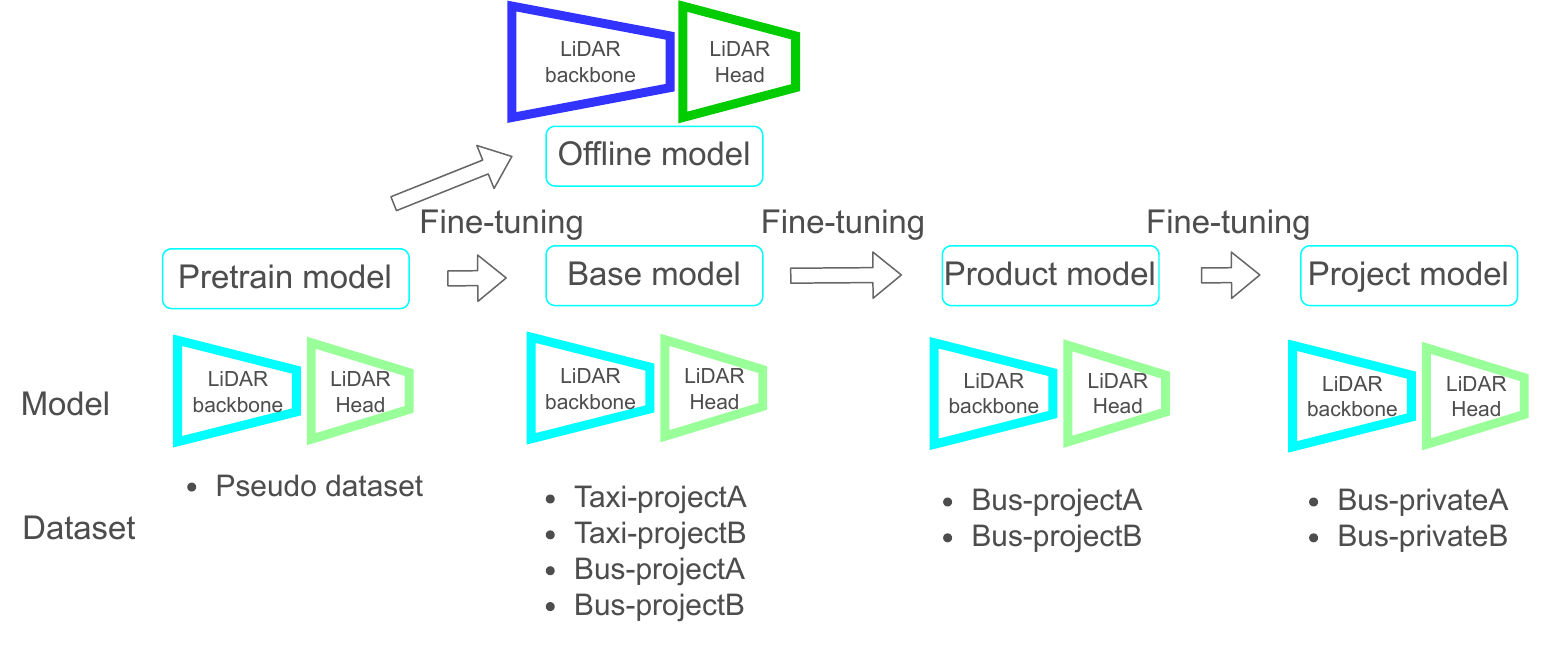}
        \caption{Model types for fine-tuning pipelines in LiDAR-only 3D object detection.}
        \label{model-tuning}
    \end{center}
\end{figure*}

As illustrated in \cref{model-examples}, we provide a variety of 3D object detection models tailored for different applications.
These include Camera-LiDAR fusion models, camera-only models, and offline models used for data mining rather than real-time deployment in vehicles.
To support diverse vehicle types and project scopes, we introduce a unified dataset format called T4dataset.
It is built upon the nuScenes format and extended to accommodate both open and private projects, as detailed in \cref{appendix-t4dataset}.
Open projects allow their data to be used across multiple training tasks, while private projects are restricted to internal use and isolated from cross-project training.
Within this framework, we define five stages of model deployment-illustrated in \cref{model-tuning} using the example of LiDAR-only 3D object detection.
As models move downstream in the pipeline, they become increasingly fine-tuned for specific vehicle types and operational environments, ensuring robust and high-performance behavior in real-world scenarios.

We first construct ``pretrain model'' from scratch, primarily trained on pseudo-labeled datasets.
This model serves as a foundation for enhancing generalization performance.

Second, we develop the ``base model'' by fine-tuning the pretrain model.
The base model is designed to be broadly applicable across various projects.
It is generally fine-tuned using the entire T4dataset, which is derived from the pretrain model.

\begin{figure*}[t]
    \begin{center}
        \includegraphics[width=0.9\linewidth]{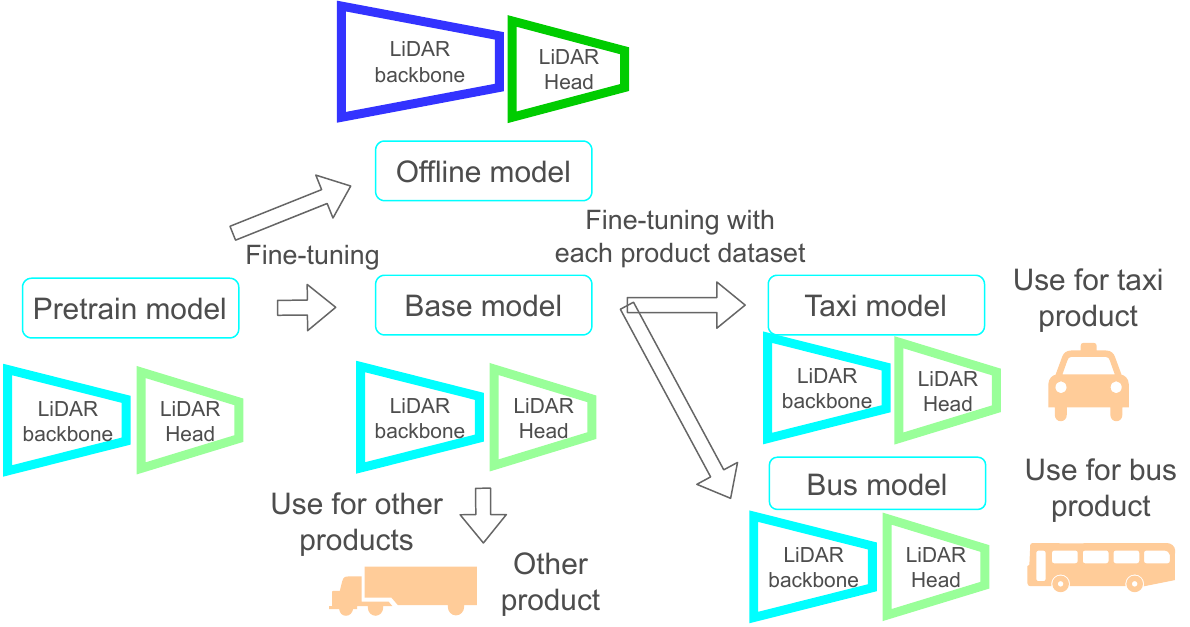}
        \caption{Usage of model.}
        \label{model-usage}
    \end{center}
\end{figure*}

Third, we construct the ``product model'' by further fine-tuning the base model.
The product model is customized for specific reference designs—such as those for taxis or buses—and is optimized for the particular sensor configurations required for deployment.
This model is fine-tuned using a subset of the T4dataset that matches the sensor configuration of the target application.
As illustrated in \cref{model-usage}, we prepare product models for taxi and bus projects, while the base model is used for other projects.

If the performance of a product model is insufficient in certain scenarios, a ``project model'' can be employed for specific projects.
The project model is tailored to individual domains and is trained using pseudo-labels generated by the offline model.
In some cases, the project model relies on proprietary or privacy-sensitive datasets that cannot be shared across projects.
Unlike other models in the AWML framework, the project model is not managed by AWML itself.
Instead, AWML only provides an interface for its integration, and the implementation and maintenance of project models are left to each respective project team.

Additionally, the ``offline model'' is employed for non-real-time processes such as pseudo-label generation, and is not intended for deployment in real-time autonomous driving systems.
The offline model is a LiDAR-only architecture specifically designed for 3D object detection, with a focus on enhancing generalization performance.
It is typically trained using the complete set of available datasets to maximize robustness and effectiveness in generating high-quality pseudo-labels.

\begin{figure*}[t]
    \begin{center}
        \includegraphics[width=0.9\linewidth]{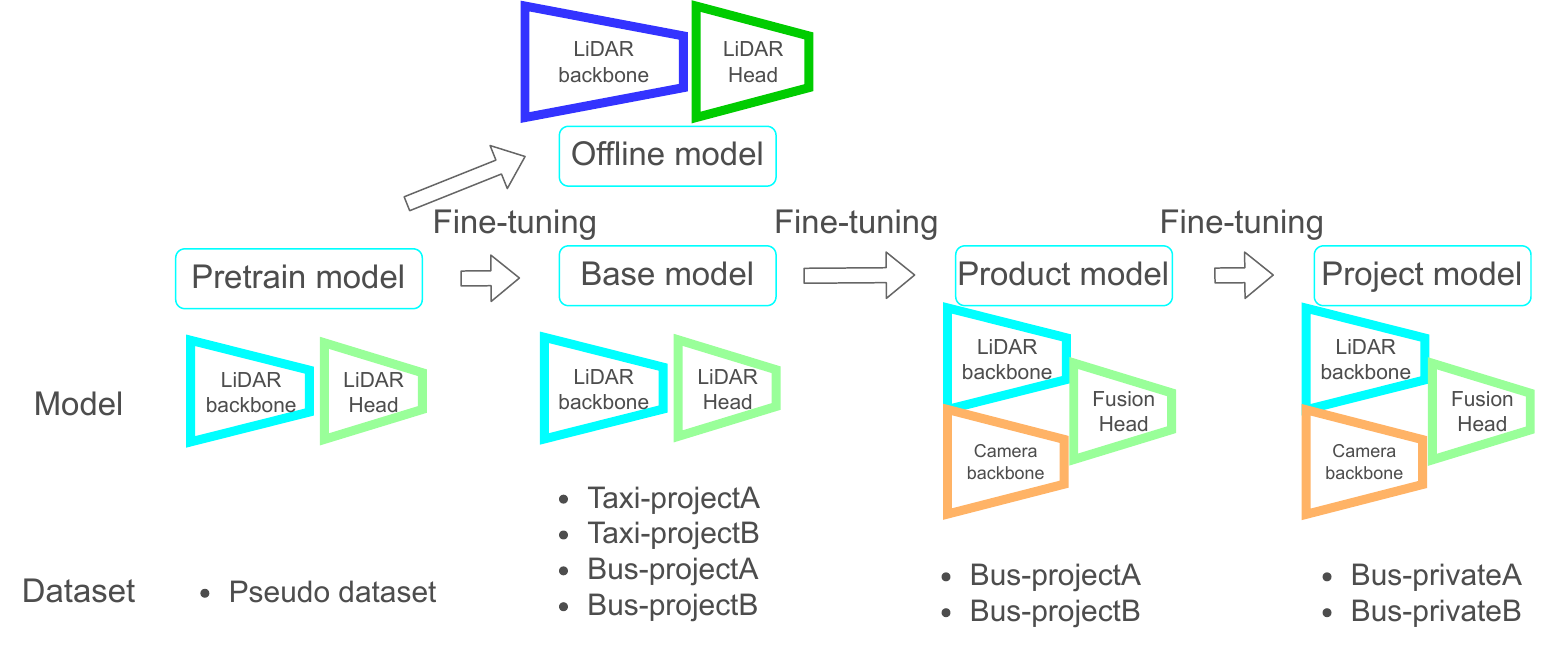}
        \caption{Model types for fine-tuning pipelines in Camera-LiDAR-Fusion 3D object detection.}
        \label{model-tuning-fusion}
    \end{center}
\end{figure*}

\cref{model-tuning-fusion} illustrates the tuning flow for Camera-LiDAR Fusion 3D object detection.
Unlike the LiDAR-only models, the product model within the Camera-LiDAR fusion architecture can be fine-tuned from the base model trained using the LiDAR-only model.

\subsection{Model Management}
\label{method-model-management}

We manage the version of ML model in AWML.
We name it as ``algorithm name + model name/version.''
For example, we use as following.

\begin{itemize}
  \item The base model of ``BEVFusion-L base/1.2.''
  \item The base model of ``BEVFusion-offline base/2.4.''
  \item The base model of ``CenterPoint-nearby base/3.1.''
  \item The product model of ``BEVFusion-L taxi/1.2.2.''
  \item The product model of ``CenterPoint bus/1.2.1.''
  \item The project model of ``CenterPoint bus/1.2.3-odaiba.2.''
\end{itemize}

\textbf{Algorithm Name.}
The algorithm name is represented as a string, such as ``CenterPoint'' or ``BEVFusion.''
Some algorithm names include modality-specific suffixes.
For example, ``BEVFusion-L'' denotes a BEVFusion model that uses LiDAR point cloud inputs, while ``BEVFusion-CL'' indicates a model utilizing both camera and LiDAR inputs.
Algorithm names can also be extended to reflect specialized variants designed for particular tasks.
For instance, ``CenterPoint-offline'' refers to a model tailored for auto-labeling, whereas ``CenterPoint-nearby'' is optimized for detecting nearby objects such as pedestrians and bicycles.

\textbf{Model Name.}
The model name follows an enumeration-based format consisting of one of ``pretrain,'' ``base,'' or a product-specific identifier.
Currently, product-specific names are derived from vehicle identifiers, which serve as references for different deployment contexts.

\textbf{Version.}
Versioning adheres to a semantic versioning scheme, typically using integer-based notation.
In certain cases, the version identifier may include additional string-based qualifiers.
Further details regarding the versioning methodology are provided in \cref{method-model-versioning}.


\subsection{Model Versioning}
\label{method-model-versioning}

\begin{figure*}[t]
    \begin{center}
        \includegraphics[width=0.9\linewidth]{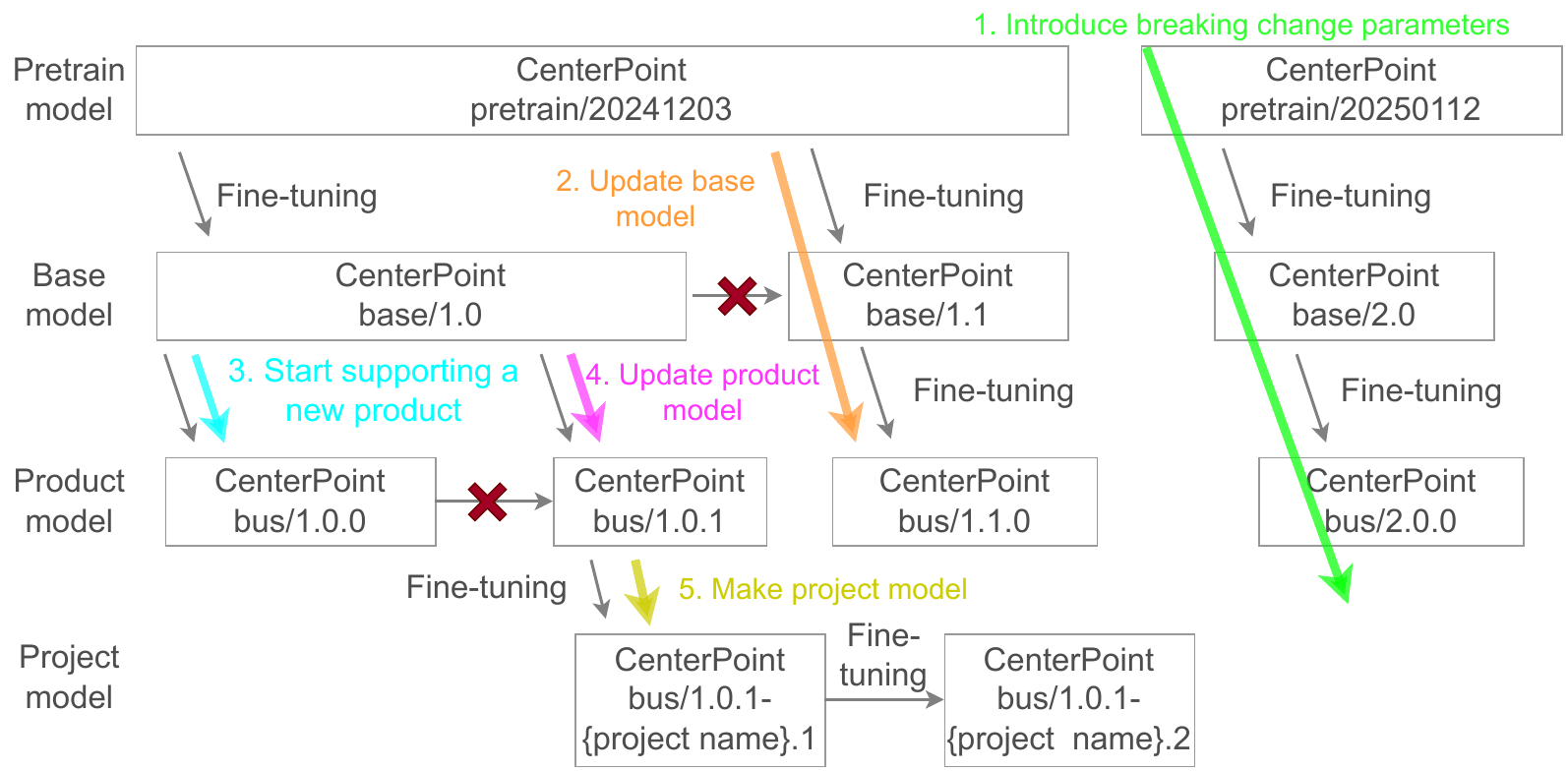}
        \caption{Model release strategy.}
        \label{model-release}
    \end{center}
\end{figure*}

We define the version of model, as shown in \cref{model-release}.

\textbf{Pretrain model: ``pretrain/\{date\}.''}
Pretrain model is an optional model, so its versioning can be skipped when managing model versions.

We prepare the pretrain model with pseudo T4dataset to increase generalization performance.
Pseudo T4dataset contains various vehicle types, sensor configurations, and kinds of LiDAR.
We aim to adapt to various sensor configurations by using a pretrain model, which is trained by various pseudo T4dataset.
As version definition, ``\{date\}'' represent the date that the model was created.
We do not use versioning and manage the model by the document which describes the used config.
As an example, we name ``CenterPoint pretrain/20241203'', which was trained on December 3rd, 2024.
To update the version of pretrain model from from ``pretrain/\{date\}'' to ``pretrain/\{next\_date\}.''

We prepare the pretrain model using a pseudo T4dataset to enhance generalization performance.
Pseudo T4dataset contains various vehicle types, sensor configurations, and different LiDAR sensors.
By training on diverse pseudo T4datasets, the pretrain model is designed to adapt to various sensor configurations.
The ``\{date\}'' field represents the creation date of the model.
Unlike other models, pretrain models are not versioned but are instead documented with details about their training configurations.
For example, ``CenterPoint pretrain/20241203'' indicates a pretrain CenterPoint model trained on December 3rd, 2024.
To update a pretrain model, the version is replaced from ``pretrain/\{date\}'' to ``pretrain/\{next\_date\}'' according to its creation date.

\textbf{Base model: ``base/X.Y.''}
When creating a new model, it should start from the base model.
``X'' represents the major version for Autoware and manages parameters related to ROS packages.
A change in the major version indicates that ROS software developers need to review and integrate the system accordingly.
If the major version remains unchanged, the model can be used with the same ROS packages.
Major version zero (``0.Y'') is designated for initial development, where breaking changes are acceptable.
``Y'' represents the version of the base model and tracks modifications in the training configuration.
This includes changes in training parameters, the dataset used, and the pretrain model.
For example, ``CenterPoint base/1.2'' indicates that the model follows version 1 of the ROS parameter configuration and has undergone two dataset updates.

The criteria for updating from ``base/X.Y'' to ``base/(X+1).0'' require changes in ROS-related parameters.
For example, if detection range configuration or additional input features affect both training parameters and ROS parameters, both the Autoware-ML configuration and ROS parameters must be updated, leading to an increment in ``X.''
The criteria for updating from ``base/X.Y'' to ``base/X.(Y+1)'' apply when the model is retrained without modifying ROS parameters.
For instance, if the training dataset changes, version ``Y'' is updated.
Similarly, if the pretrain model changes but ROS-related parameters remain the same, version ``Y'' is incremented.
For Autoware users, as long as version ``X'' remains unchanged, the updated model can be used with the same version of the ROS package.

\textbf{Product model: ``\{product\_name\}/X.Y.Z.''}
Product model is an optional model for deployment, so this section can be skipped when managing model versioning.
If it is necessary to improve perception performance for a specific product (i.e., a particular vehicle), a product model should be prepared.
The version definition consists of the following components:
``\{product\_name\}'' represents the name of the product.
``X.Y'' corresponds to the version of the base model.
``Z'' represents the version of the product model.
For example, ``CenterPoint bus/1.2.0'' indicates the initial release of a model fine-tuned from ``CenterPoint base/1.2'' for the bus product.
Similarly, ``CenterPoint bus/1.2.3'' refers to a model fine-tuned from ``CenterPoint base/1.2'' for the bus product, updated for the third time.

\textbf{Project model: ``\{product\_name\}/X.Y.Z-\{project\_version\}.''}
If an issue arises in the product model and a temporary fix is needed using a pseudo dataset, a project model is released.
The performance of the project model does not differ significantly from the product model.
It is important to note that the project model is a temporary solution, and the official release will be the next version of the product model, retrained with annotated data for the issue scene.
Because of this, Pseudo-T4dataset and project models are not managed by AWML and are managed by projects.
The version definition consists of the following components: ``\{product\_name\}/X.Y.Z'' represents the version of the product model, and ``\{project\_version\}'' denotes the version of the project model.
Semantic versioning pre-release notation is used for the project version.
However, unlike standard semantic versioning where ``X.Y.Z'' $>$ ``X.Y.Z-\{project\_version\}'', here ``X.Y.Z'' $<$ ``X.Y.Z-\{project\_version\}'' to indicate that ``\{project\_version\}'' is a newer model version.
For example, ``CenterPoint bus/1.2.3-odaiba.2'' represents a project model fine-tuned from ``CenterPoint bus/1.2.3.''
As a criterion for version updates in ``\{product\_name\}/X.Y.Z-\{project\_version\}'', an update from ``CenterPoint bus/1.2.3-odaiba.2'' to ``CenterPoint bus/1.2.3-odaiba.3'' signifies an improved project model version.

\subsection{Strategy for Fine-Tuning}
\label{method-fine-tuning}

\textbf{Introduce breaking change parameters.}
``1. Introduce breaking change parameters'' in \cref{model-release} shows the introduction of a breaking change for the ROS package.
In this release, we update from ``base/X.Y'' to ``base/(X+1).0'' and update from ``\{product\_name\}/X.Y.Z'' to ``\{product\_name\}/(X+1).0.0.''
This update includes changes related to ROS parameters, such as range parameters.
Since these changes require creating a new model from the ground up, both the base model and product model must be released.
Additionally, if the base model update necessitates changes to the pretrain model, the pretrain model should also be retrained accordingly.

\textbf{Update base model.}
``2. Update base model'' in \cref{model-release} shows the model update by adding an additional database dataset.
In this release, we update from ``base/X.Y'' to ``base/X.(Y+1)'' and update from ``\{product\_name\}/X.Y.Z'' to ``\{product\_name\}/X.(Y+1).0.''
In general, every few months, we fine-tune the base model from the pretrain model and release the next version of the base model.
Note that we do not fine-tune the base model from itself, but from the pretrain model.
The reason we use all of the dataset is based on the strategy of a foundation model.
The base model is fine-tuned to adapt to a wide range of sensor configurations and driving areas.
Note that if we support two or more base models (e.g., base/1.1 and base/2.0), we either update all the models or deprecate the old versions.
We update all the base model versions: base/1.1 to base/1.2 and base/2.0 to base/2.1.
We also update all the product model versions depending on those models, if necessary: bus/1.1.3 to bus/1.2.0 and bus/2.0.0 to bus/2.1.0.

\textbf{Start supporting a new product.}
``3. Start supporting a new product'' in \cref{model-release} shows the introduction for new product by fine-tuning from the base model.
In this release, we start making the product model from ``base/X.Y'' to ``\{product\_name\}/X.Y.0''
For example, when we start releasing CenterPoint for the product of Bus, we fine-tune from ``CenterPoint base/X.Y'' and release the product model as ``CenterPoint bus/X.Y.0.''

\textbf{Update product model.}
``4. Update product model'' in \cref{model-release} shows the model update by adding product dataset.
In this release, we update from ``\{product\_name\}/X.Y.Z'' to ``\{product\_name\}/X.(Y+1).0.''
When a new annotated T4dataset is added, we release the product to fine-tune the base model.
Note that updating the product model will NOT trigger the project model, as they are temporary releases or release candidates.
Also, we do not fine-tune the product model from the product model, but from the base model.

\textbf{Make project model.}
``5. Make project model'' in \cref{model-release} shows the introduction for new project.
In this release, we create from ``\{product\_name\}/X.Y.Z'' to ``\{product\_name\}/X.Y.Z-\{project\_version\}.''
If there is an issue from a particular project, we create a project model to deploy a band-aid model for that project.
Fine-tuning is done from product model using the private dataset or Pseudo-T4dataset, which is created by the offline model.
For example, the project model ``CenterPoint bus/1.2.3-odaiba.1'' is fine-tuned from ``CenterPoint bus/1.2.3.''
AWML does not manage project models and Pseudo-T4dataset, and it is acceptable to retrain from a project model.

\subsection{Development of New Algorithm}
\label{method-new-algorithm}

\begin{figure*}[t]
    \begin{center}
        \includegraphics[width=0.99\linewidth]{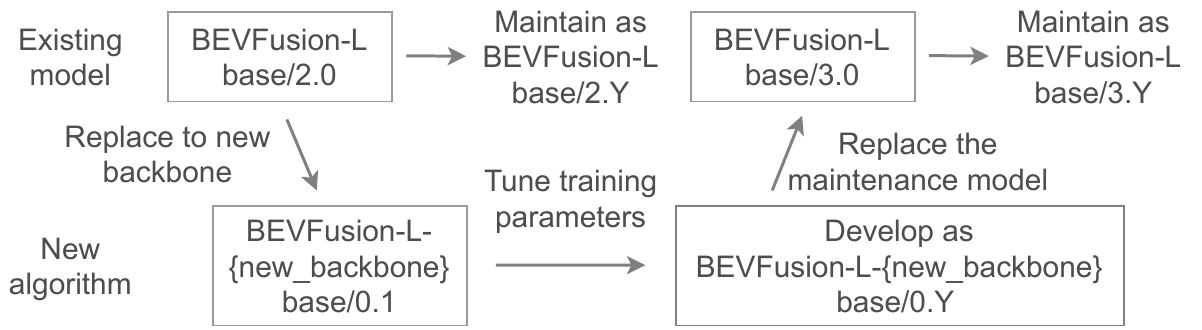}
        \caption{Introduction of new algorithm.}
        \label{new-algorithm}
    \end{center}
\end{figure*}

If a new algorithm is to be introduced, such as replacing the backbone or head, new models should be created as part of the new algorithm, as shown in \cref{new-algorithm}.
Rather than making disruptive changes to an existing model, development should begin as a separate model during the experimental phase.
\cref{new-algorithm} provides an example of introducing a new backbone to BEVFusion-L.
In major version zero (``0.Y''), initial development is carried out, allowing breaking changes.
This strategy maintains two branches.
First branch is a maintenance branch for existing models to ensure stability for operation engineers.
Second branch is a development branch for new algorithms to explore performance improvements.
After completing the core algorithm development, parameter tuning and fine-tuning are typically conducted for each specific product.
Once a stable model suitable for production deployment is achieved, the major version is incremented.

\subsection{Strategy for Product-Release Model}
\label{method-model-release}

\begin{figure*}[t]
    \begin{center}
        \includegraphics[width=0.9\linewidth]{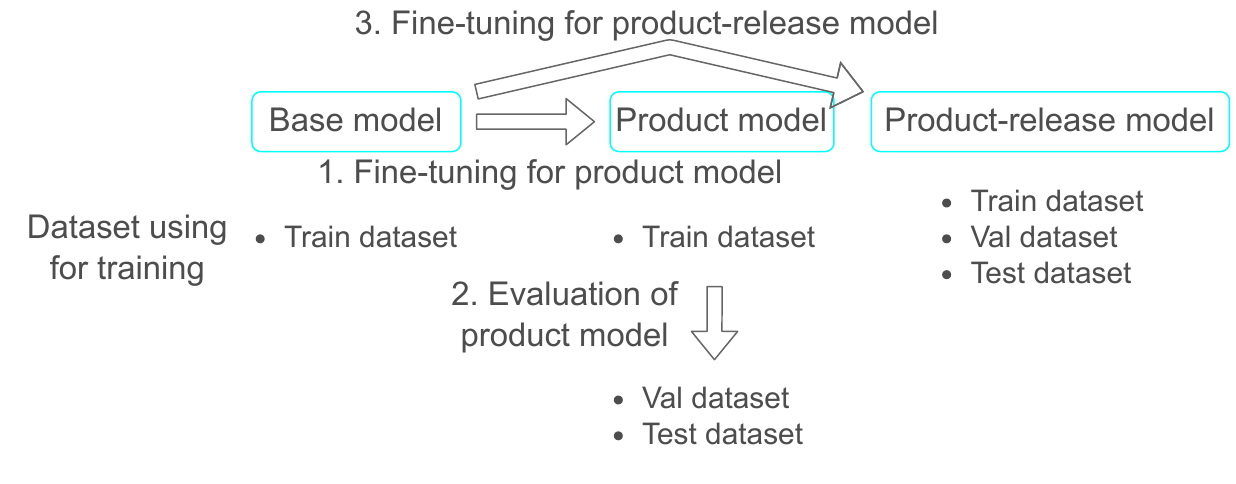}
        \caption{Fine-tuning strategy.}
        \label{release-model}
    \end{center}
\end{figure*}

In typical machine learning workflows, datasets are divided into train, validation, and test splits.
This separation is essential to detect overfitting, assess generalization performance (i.e., performance on unseen data), and provide a reliable basis for tuning and comparison.
Accordingly, only the training dataset is used for model training.
However, in product-level deployment, especially in domains like autonomous driving, the goal shifts toward maximizing performance in real-world environments.
In such contexts, data collection and annotation are often costly and time-consuming, and acquiring additional labeled data can be challenging.
Therefore, there is a strong motivation to utilize as much available data as possible for model training.
In particular, during early industrial phases where labeled data is few, if generalization performance has already been sufficiently evaluated, or if visual validation without labels is considered, it can be a viable strategy to train on the entire dataset to prioritize final model accuracy.

To address this need, AWML introduces an option for generating product-level models, called the product-release model.
This process is illustrated in \cref{release-model}, and consists of the following steps.
Fine-tuning a base model using only the training dataset to create a product model (step 1).
Evaluating the product model using standard validation and/or test data (step 2).
Fine-tuning a new product-release model from the original base model using the full dataset (including train, val, and test splits), based on the best-performing configuration found in step 2 (step 3).
This model is then released as the product-release model.
For example, we call ``CenterPoint bus/1.1.1-release'' as product-release model.
By doing so, the product-release model benefits from learning on the full dataset-including what was previously reserved for validation and testing.

Importantly, AWML does not provide release models for the base model.
This is a deliberate design choice: in our fine-tuning strategy, the base model serves as the source for all downstream fine-tuning, and using a release model at this stage would risk data leakage.

\section{Experiment}

\subsection{Setting}
\label{exp-setting}

To evaluate the pipeline from training to deploy for ROS 2 environment, we make internal dataset in Japanese area with different sensor configuration.
We prepare two T4datasets, Taxi dataset and Bus dataset.
Taxi dataset contains about 21k frames and Bus dataset contains about 13k frames.
Our T4dataset contain 5 classes, car, truck, bus, bicycle, and pedestrian.

In \cref{exp-3ddet}, we evaluate 3D object detection models.
In our experiment, we set experimental condition as same as nuScenes dataset.
We use mAP metrics and its threshold of center distance is \SI{0.5}{m}, \SI{1.0}{m}, \SI{1.5}{m}, \SI{2.0}{m}.
We evaluate CenterPoint\cite{yin2021center} ,PillarNeSt\cite{jia2023adriver}, BEVFusion-L, and BEVFusion-CL \cite{liu2022bevfusion} as 3D object detection models.
In our models, our CenterPoint use pillar-based encoder \cite{Lang2018PointPillarsFE} and SECOND backbone \cite{Yan2018SECONDSE}.
Our PillarNeSt use pillar-based encoder and PillarNeSt backbone based on CenterPoint.
BEVFusion-L is LiDAR-only model using sparse convolution encoder.
BEVFusion-CL is Camera-LiDAR fusion model.

In \cref{exp-2ddet}, we evaluate 2D detection for traffic light using T4dataset for traffic light recognition.
In this experiment, we use YOLOX\_opt, which we optimize the layer number for our system from YOLOX \cite{Ge2021YOLOXEY}, to 2D detection.
We use MobileNetv2 \cite{Sandler2018MobileNetV2IR} to 2D classification.

\subsection{3D Object Detection}
\label{exp-3ddet}

\begin{table}[t]
  \centering
  \caption{
    Benchmark of 3D object detection for test dataset of all T4dataset.
    Evaluation range is \SI{120}{m}.
    ``Tru.'' represents truck, ``Bic.'' represents bicycle, and ``Ped.'' represents pedestrian.
  }
  \begin{tabular}{l|c|p{14pt}p{14pt}p{14pt}p{14pt}p{14pt}}
      Algorithm     & mAP & Car & Tru. & Bus & Bic. & Ped. \\ \hline
      CenterPoint   & 64.4 & 75.0 & 50.7 & 78.1 & 53.2 & 64.8 \\
      PillarNeSt    & 68.6 & 77.9 & 58.6 & 80.9 & 53.2 & 64.8 \\
      BEVFusion-L   & 63.7 & 72.3 & 58.8 & 70.5 & 70.2 & 56.3 \\
      BEVFusion-CL  & 66.9 & 80.6 & 65.7 & 59.7 & 66.8 & 60.2 \\ \hline
  \end{tabular}
  \label{exp-3ddet-1}
\end{table}

\begin{table}[t]
  \centering
  \caption{
    Benchmark of 3D object detection for test dataset of Taxi T4dataset.
    Evaluation range is \SI{120}{m}.
  }
  \begin{tabular}{l|c|p{14pt}p{14pt}p{14pt}p{14pt}p{14pt}}
      Algorithm & mAP & Car & Tru. & Bus & Bic. & Ped. \\ \hline
      CenterPoint   & 63.3 & 74.8 & 53.3 & 66.4 & 56.3 & 65.5 \\
      PillarNeSt    & 67.8 & 77.1 & 57.4 & 73.9 & 61.5 & 69.1 \\
      BEVFusion-L   & 68.2 & 75.6 & 63.2 & 71.5 & 70.0 & 60.6 \\
      BEVFusion-CL  & 68.5 & 76.0 & 64.9 & 68.6 & 73.2 & 60.1 \\ \hline
  \end{tabular}
  \label{exp-3ddet-2}
\end{table}

\begin{table}[t]
  \centering
  \caption{
    Benchmark of 3D object detection for test dataset of Bus T4dataset.
    Evaluation range is \SI{120}{m}.
  }
  \begin{tabular}{l|c|p{14pt}p{14pt}p{14pt}p{14pt}p{14pt}}
      Algorithm & mAP & Car & Tru. & Bus & Bic. & Ped. \\ \hline
      CenterPoint   & 64.5 & 75.0 & 45.5 & 86.6 & 51.2 & 63.9 \\
      PillarNeSt    & 68.6 & 78.5 & 55.5 & 86.4 & 54.3 & 69.3 \\
      BEVFusion-L   & 63.8 & 70.7 & 55.2 & 69.9 & 70.3 & 52.9 \\
      BEVFusion-CL  & 63.3 & 70.8 & 55.7 & 64.8 & 69.9 & 55.9 \\ \hline
  \end{tabular}
  \label{exp-3ddet-3}
\end{table}

\begin{table}[t]
  \centering
  \caption{
    The evaluation of performance.
    ``Time (P)'' represents time on PyTorch environment.
    ``Time (R)'' represents time on C++ TensorRT with ROS 2 environment.
    In this experiment, we use RTX3060 GPU.
  }
  \begin{tabular}{l|c|cc}
      Model & Range & Time (P) & Time (R) \\ \hline
      CenterPoint   & \SI{120}{m} & \SI{198.2}{ms}  & \SI{32.1}{ms} \\
      PillarNeSt    & \SI{120}{m} & \SI{299.8}{ms}  & \SI{38.5}{ms} \\
      BEVFusion-L   & \SI{120}{m} & \SI{558.2}{ms}  & \SI{29.1}{ms} \\
      BEVFusion-CL  & \SI{120}{m} & \SI{1014.2}{ms} & \SI{76.1}{ms} \\
  \end{tabular}
  \label{exp-3ddet-6}
\end{table}

\begin{table}[t]
  \centering
  \caption{
    Benchmark of offline 3D object detection for test dataset of all T4datasets.
    Evaluation range is \SI{120}{m}.
  }
  \begin{tabular}{l|c|p{14pt}p{14pt}p{14pt}p{14pt}p{14pt}}
      Algorithm & mAP & Car & Tru. & Bus & Bic. & Ped. \\ \hline
      BEVFusion-L  & 72.2 & 81.2 & 60.5 & 72.2 & 73.0 & 73.9 \\
      BEVFusion-CL & 73.7 & 81.2 & 61.1 & 84.5 & 70.2 & 71.6 \\ \hline
  \end{tabular}
  \label{exp-3ddet-4}
\end{table}

\begin{table}[t]
  \centering
  \caption{
    Fine-tuning of 3D object detection for test dataset of Bus T4dataset.
    Evaluation range is \SI{120}{m}.
    ``FT'' represents fine-tuning as product model.
  }
  \begin{tabular}{l|c|p{14pt}p{14pt}p{14pt}p{14pt}p{14pt}}
      Algorithm & mAP & Car & Tru. & Bus & Bic. & Ped. \\ \hline
      CenterPoint & 64.5 & 75.0 & 45.5  & 86.6 & 51.2 & 63.9 \\
      w/FT        & 65.4 & 75.1 & 46.4  & 90.4 & 52.6 & 62.6 \\ \hline
  \end{tabular}
  \label{exp-3ddet-5}
\end{table}

We train 3D object detection models by all dataset with Taxi dataset and Bus dataset.
We conducted extensive evaluations on 3D object detection using the T4dataset, covering different Taxi dataset and Bus datasets.
The evaluation results are presented from \cref{exp-3ddet-1} to \cref{exp-3ddet-6}.

\cref{exp-3ddet-1} shows the benchmark results for test dataset of all T4dataset.
Among the evaluated methods, PillarNeSt achieves the highest overall mAP of 68.6, showing strong performance particularly for large objects such as buses and trucks.
BEVFusion-CL also demonstrates competitive results, especially on the bicycle and pedestrian classes.
This highlights the effectiveness of multi-modal and contrastive approaches in improving detection performance across diverse object types.
According to previous studies, BEVFusion-L typically outperforms CenterPoint on the nuScenes dataset; however, we observe the opposite trend on the T4 dataset.
This difference may be attributed to variations in sensor configurations and object classes between T4 and nuScenes.
In addition, nuScenes evaluates detection performance within a range of 50 meters, whereas the T4 dataset uses a range of 120 meters.
This substantial difference in evaluation distance likely leads to different requirements for detection capabilities, which may also explain the observed performance gap.

\cref{exp-3ddet-2} and \cref{exp-3ddet-3} present the benchmark results of 3D object detection on the test datasets of the Taxi and Bus splits of the T4dataset, respectively.
On the Taxi dataset, BEVFusion variants generally outperform CenterPoint and PillarNeSt, with BEVFusion-CL achieving the highest mAP of 68.5.
The improvements are particularly notable in the Truck and Bicycle categories.
On the Bus dataset, PillarNeSt achieves the best overall performance with an mAP of 68.6, showing strong results across most object classes.
While BEVFusion variants maintain competitive scores in certain classes such as Bicycle, their overall mAP is lower compared to the Taxi dataset.
These results indicate that detection performance varies depending on sensor configurations and object distributions, underscoring the importance of evaluating models across diverse environments.

\cref{exp-3ddet-6} presents the inference time comparison of four models evaluated on an RTX3060 GPU.
We report latency under two execution environments: PyTorch (denoted as ``Time (P)'') and C++ TensorRT with ROS 2 (denoted as ``Time (R)'').
As expected, inference time is significantly reduced in the optimized ROS 2 + TensorRT environment compared to the PyTorch baseline.
For instance, BEVFusion-L demonstrates a considerable speedup from 558.2 ms in PyTorch to 29.1 ms in TensorRT, while CenterPoint reduces from 198.2 ms to 32.1 ms.
Notably, BEVFusion-CL, while achieving strong detection performance, incurs higher latency under both environments, reflecting the computational cost.

\cref{exp-3ddet-4} shows the benchmark for test dataset of all T4datasets about offline models.
BEVFusion-CL outperforms BEVFusion-L in overall mAP (73.7 vs. 72.2), and shows significant gains particularly in the bus category (84.5 vs. 72.2).
Compared to online models, offline models improve for small objects, bicycle and pedestrian.

\cref{exp-3ddet-5} shows the benchmark for test dataset of Bus dataset to evaluate the effectiveness of fine-tuning pipeline.
After fine-tuning (denoted as ``w/FT''), the overall mAP improved from 64.5 to 65.4.
The Bus category showed a performance gain, increasing from 86.6 to 90.4 in AP.
This result suggests that fine-tuning for product model can effectively enhance the model's performance in target product.

\subsection{2D Object Detection}
\label{exp-2ddet}

\begin{table}[t]
  \centering
  \caption{
    2D detection for traffic light.
  }
  \begin{tabular}{l|c}
      Metric & YOLOX\_opt \\ \hline
      mAP    & 0.3562 \\
      AP50   & 0.4790 \\
      AP60   & 0.4730 \\
      AP70   & 0.4480 \\
      AP80   & 0.3250 \\
      AP90   & 0.0560 \\
  \end{tabular}\label{exp-2ddet-1}
\end{table}

\begin{table}[t]
  \centering
  \caption{
    2D classification for traffic light.
  }
  \begin{tabular}{l|c}
      Metric & MobileNetv2 \\ \hline
      Precision (Top-1) & 69.75 \\
      Recall (Top-1)    & 68.09 \\
      F1-score (Top-1)  & 68.85 \\
  \end{tabular}
  \begin{tabular}{lc}
  \end{tabular}\label{exp-2ddet-2}
\end{table}

\begin{table}[t]
  \centering
  \caption{
    The evaluation of performance.
    ``Time (PyTorch)'' represents performance on A100 GPU with pytorch environment with batchsize = 1.
    ``Time (ROS 2)'' represents performance on RTX3090 GPU with C++ TensorRT and ROS 2 environment with batchsize = 1.
  }
  \begin{tabular}{l|ccc}
      Model & Time (PyTorch) & Time (ROS 2) \\ \hline
      YOLOX\_opt & \SI{19.4}{ms}  & \SI{1.9}{ms} \\
      MobileNetv2 & \SI{9.8}{ms} & \SI{1.7}{ms} \\
  \end{tabular}
  \label{exp-2ddet-3}
\end{table}

We conducted experiments to evaluate the performance of our models for 2D traffic light detection and classification.
The results are presented in Tables \cref{exp-2ddet-1}, \cref{exp-2ddet-2}, and \cref{exp-2ddet-3}.
\cref{exp-2ddet-1} presents the 2D object detection performance of the YOLOX\_opt model.
The model achieves an overall mAP of 0.3562, with a notable drop in performance at higher IoU thresholds such as AP90.
\cref{exp-2ddet-2} presents the 2D classification performance of the MobileNetv2 model.
\cref{exp-2ddet-3} compares the inference speed of both models in two different environments, PyTorch and ROS 2.
The YOLOX\_opt model achieves an inference time of \SI{19.4}{ms} in the PyTorch environment and \SI{1.9}{ms} in the ROS 2 environment.
Similarly, the MobileNetv2 model achieves \SI{9.8}{ms} in PyTorch and \SI{1.7}{ms} in ROS 2.
These results highlight the efficiency of the models when deployed in different real-world scenarios.

\section{Conclusion}
\label{sec:intro}

In this paper, we present AWML, a machine learning framework designed for autonomous driving and robotics MLOps.
AWML supports not only the deployment of trained models to robotic systems but also incorporates an active learning framework, including auto-labeling, semi-auto-labeling, and data mining techniques.
We also conduct extensive benchmarking on baseline algorithms for 3D object detection, evaluate fine-tuning strategies, validate deployment in ROS-based environments.

For future work, we plan to extend AWML to support 2D and 3D semantic segmentation with integrated active learning and ROS deployment.
We also aim to further improve the performance of 3D object detection by leveraging active learning within this framework.
Ultimately, we envision AWML as a bridge between the computer vision and robotics communities, between supervised and active learning, and between research and real-world deployment, contributing to the advancement of autonomous driving technologies.

{
    \small
    \bibliographystyle{ieeenat_fullname}
    \bibliography{main}
}

\clearpage
\setcounter{page}{1}
\maketitlesupplementary

\section{Appendix}
\subsection{Contribution}
\label{appendix-contribution}

\begin{itemize}
  \item Tanaka: Design whole architecture of AWML.
  \item Tanaka, Samrat, Kok, Amadeusz, Kenzo, Minoda, Tomie, Uetake: Main contributor for AWML.
  \item Zhang, Minoda: Contributor of constructing T4dataset.
  \item Minoda, Yamashita, Horibe: Management of research and development.
\end{itemize}

\subsection{Software architecture of AWML}
\label{appendix-awml}

\begin{figure}[b]
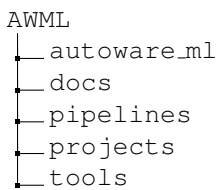

  \dirtree{%
  .1 AWML.
  .2 autoware\_ml.
  .2 docs.
  .2 pipelines.
  .2 projects.
  .2 tools.
  }
  \caption{The whole structure of AWML}
  \label{AWML-structure}
\end{figure}

The framework of AWML consists of the library of autoware\_ml, pipelines, projects, and tools as shown in \cref{AWML-structure}.

\textbf{\textcolor{blue}{autoware\_ml/}:}
The directory of ``autoware\_ml/'' provide library for AWML.
This directory can be used as library and this directory doesn't depend on other directories.
The libraries of ``autoware\_ml/{task name}'' provide the core library for each task.
It may contains dataset loader and metrics for T4dataset.
The directory of ``autoware\_ml/configs'' contains the config files are used commonly for each projects.
The config file like ``autoware\_ml/configs/detection3d/db\_jpntaxi\_v1.yaml'' defines dataset ids of T4dataset.

\begin{figure}[t]
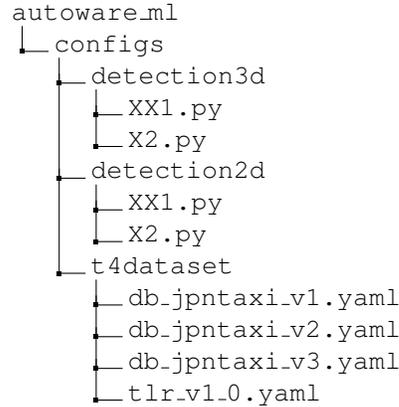

  \dirtree{%
  .1 autoware\_ml.
  .2 configs.
  .3 detection3d.
  .4 XX1.py.
  .4 X2.py.
  .3 detection2d.
  .4 XX1.py.
  .4 X2.py.
  .3 t4dataset.
  .4 db\_jpntaxi\_v1.yaml.
  .4 db\_jpntaxi\_v2.yaml.
  .4 db\_jpntaxi\_v3.yaml.
  .4 tlr\_v1\_0.yaml.
  }
  \caption{The structure of autoware\_ml}
\end{figure}

\textbf{\textcolor{blue}{docs/}:}
The directory of ``docs/'' is design documents for AWML.
The target of documents is a designer of whole ML pipeline system and developers of AWML core library.

\textbf{\textcolor{blue}{pipelines/}:}
The directory of ``pipelines/'' manages the pipelines that consist of ``tools'' .
This directory can depend on ``/autoware\_ml'' , ``projects'' , ``/tools'' , and other ``/pipelines'' .
Each pipeline has ``README.md'' , a process document to use when you ask someone else to do the work.
The target of ``README.md'' is a user of AWML.

\textbf{\textcolor{blue}{projects/}:}
The directory of ``projects/'' manages the model for each tasks.
This directory can depend on ``/autoware\_ml'' and other ``projects'' .
Each project has ``README.md'' for users.
The target of ``README.md'' is a user of AWML.

\begin{figure}[t]
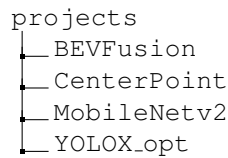

  \dirtree{%
  .1 projects.
  .2 BEVFusion.
  .2 CenterPoint.
  .2 MobileNetv2.
  .2 YOLOX\_opt.
  }
  \caption{The structure of projects}
  \label{autoware_ml_project}
\end{figure}

\textbf{\textcolor{blue}{tools/}:}
The directory of ``tools/'' manages the tools for each tasks.
'' tools/'' scripts are abstracted. For example, ``tools/detection3d'' can be used for any 3D detection models such as CenterPoint and BEVFusion.
This directory can depend on ``/autoware\_ml'' and other ``/tools'' .
Each tool has ``README.md'' for developers.
The target of ``README.md'' is a developer of AWML.

\begin{figure}[t]
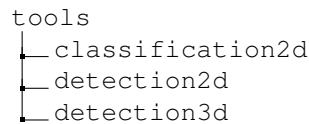

  \dirtree{%
  .1 tools.
  .2 classification2d.
  .2 detection2d.
  .2 detection3d.
  }
  \caption{The structure of tools}
  \label{autoware_ml_tool}
\end{figure}

\subsection{T4dataset}
\label{appendix-t4dataset}

\begin{figure}[t]
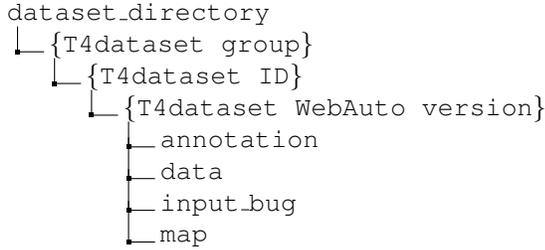

  \dirtree{%
  .1 dataset\_directory.
  .2 \{T4dataset group\}.
  .3 \{T4dataset ID\}.
  .4 \{T4dataset WebAuto version\}.
  .5 annotation.
  .5 data.
  .5 input\_bug.
  .5 map.
  }
  \caption{The structure of T4dataset}
  \label{t4dataset_structure}
\end{figure}

\textbf{T4dataset format.}
To use for training and active learning with ROS environment, we define T4dataset format based on nuScenes format.
The directory architecture is shown in \cref{t4dataset_structure}.
'' T4dataset group'' consists of the type of T4dataset and dataset version.
For example, ``db\_jpntaxi\_v1'' means the database T4dataset for the vechile of JpnTAXI and its version is 1.
'' T4dataset ID'' is the id managed in WebAuto system \cite{WebAuto} as ``70891309-ca8b-477b-905a-5156ffb3df65'' .
'' T4dataset WebAuto version'' is the version of T4dataset itself.
If we fix annotation or sensor data, we update this version.
When we make a T4dataset, we start from version 0.

\textbf{t4dev-kit.}
The software of ``t4dev-kit'' provide the API to use T4dataset.
It provides the loading, evaluation, visualization.

\textbf{tier4\_perception\_dataset.}
The software of ``tier4\_perception\_dataset'' provide the scripts from rosbag to annotated T4dataset.

\textbf{autoware\_perception\_evaluation.}
The software of ``autoware\_perception\_evaluation'' provide the API of evaluation with T4dataset.

\subsection{T4Dataset types}
\label{appendix-t4dataset-types}

\begin{figure}[t]
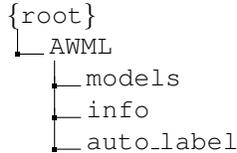

  \dirtree{%
  .1 \{root\}.
  .2 AWML.
  .3 models.
  .3 info.
  .3 auto\_label.
  }
  \caption{The structure of model zoo}
  \label{model-zoo}
\end{figure}

\begin{figure*}[t]
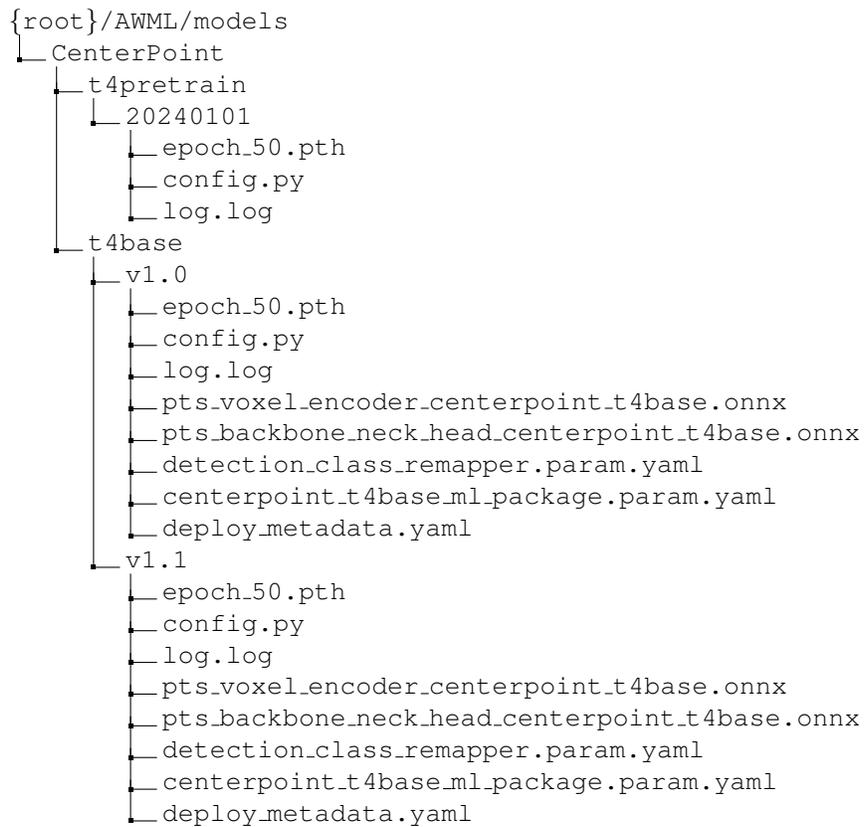

  \dirtree{%
  .1 \{root\}/AWML/models.
  .2 CenterPoint.
  .3 t4pretrain.
  .4 20240101.
  .5 epoch\_50.pth.
  .5 config.py.
  .5 log.log.
  .3 t4base.
  .4 v1.0.
  .5 epoch\_50.pth.
  .5 config.py.
  .5 log.log.
  .5 pts\_voxel\_encoder\_centerpoint\_t4base.onnx.
  .5 pts\_backbone\_neck\_head\_centerpoint\_t4base.onnx.
  .5 detection\_class\_remapper.param.yaml.
  .5 centerpoint\_t4base\_ml\_package.param.yaml.
  .5 deploy\_metadata.yaml.
  .4 v1.1.
  .5 epoch\_50.pth.
  .5 config.py.
  .5 log.log.
  .5 pts\_voxel\_encoder\_centerpoint\_t4base.onnx.
  .5 pts\_backbone\_neck\_head\_centerpoint\_t4base.onnx.
  .5 detection\_class\_remapper.param.yaml.
  .5 centerpoint\_t4base\_ml\_package.param.yaml.
  .5 deploy\_metadata.yaml.
  }
  \caption{The structure of model data}
  \label{model-zoo-model}
\end{figure*}

We divide the four types for T4dataset as following.

\textbf{Database T4dataset.}
Database T4dataset is mainly used for training a model.
We call database T4dataset as ``Database {vehicle name} vX.Y'' , ``DB {vehicle name} vX.Y'' in short.
For example, we use like ``Database JPNTAXI v1.1'' , ``DB JPNTAXI v1.1'' in short.
We manage database T4dataset in dataset configs like ``db\_jpntaxi\_v1.yaml'' (file name use only the version of X).

\textbf{Use case T4dataset.}
Use case T4dataset is mainly used for evaluation with ROS environment.
We call Use case T4dataset as ``Use case {vehicle name} vX.Y'' , ``UC {vehicle name} vX.Y'' in short.

\textbf{Non-annotated T4dataset.}
Non-annotated T4dataset is the dataset which is not annotated.
After we annotate for it, it change to database T4dataset or use case T4dataset.

\textbf{Pseudo T4dataset.}
Pseudo T4dataset is annotated to non-annotated T4dataset by auto-labeling.
Pseudo T4dataset is mainly used to train pre-training model.
We call pseudo T4dataset as ``Pseudo {vehicle name} vX.Y'' .
For example, we use like ``Pseudo JPNTAXI v1.0'' .
We manage pseudo T4dataset in dataset configs like ``pseudo\_jpntaxi\_v1.yaml'' (file name use only the version of X).
Note that AWML do not manage Pseudo T4dataset which is used for domain adaptation.

\subsection{Versioning strategy for Database T4dataset}
\label{appendix-t4dataset-version}

We manage the version of T4dataset as ``the type of T4dataset'' + ``vehicle name'' + ``vX.Y'' .
For example, we use like ``DB JPNTAXI v2.2'' , ``DB GSM8 v1.1'' , ``Pseudo J6Gen2 v1.0'' .
As the type of T4dataset, we use ``DB'' for database T4dataset, ``UC'' for use case T4dataset, and ``Pseudo'' for pseudo T4dataset.

As vehicle name, We make JPNTAXI dataset as an example
'' JPNTAXI'' is the taxi based on ``JPN TAXI'' .
It is categorized in Robo-Taxi, what we call ``XX1'' .
The sensor configuration basically consists of 1 * VLS-128 as top LiDAR, 3 * VLP-16 as side LiDARs, 6 * TIER IV C1 cameras (85deg) as cameras for object recognition, 1 * TIER IV C1 camera (85deg) as camera for traffic light recognition, 1 * TIER IV C2 camera (62deg) as camera for traffic light recognition, and 6 * continental ARS408-21 as radars.
The sensors depends on when the data was taken.

X is management classification for dataset
It is recommended to change the number depending on the location and data set creation time.

Y is the version of dataset
Upgrade the version every time a change may have a negative impact on performance for training.
For example, if we change of the way to annotation, we update the dataset and this version.
If we add new dataset, we update this version.
If we update the T4 format, we update the dataset and this version.

\subsection{Model zoo}
\label{appendix-model-zoo}

We use AWS S3 storage to manage models and intermediate product.
\cref{model-zoo} shows the architecture of directory.
We use the model registered in S3 storage for Fine-tuning like model zoo.
In AWML, we can use URL path instead of local path as MMLab libraries.
Model data to manage in S3 is shown in \cref{model-zoo-model}.

\end{document}